\documentclass{article} % For LaTeX2e
\usepackage{iclr2027_conference,times}

\usepackage{amsmath,amsfonts,bm}

\def\eqref#1{equation~\ref{#1}}
\def\1{\bm{1}}

\DeclareMathAlphabet{\mathsfit}{\encodingdefault}{\sfdefault}{m}{sl}
\SetMathAlphabet{\mathsfit}{bold}{\encodingdefault}{\sfdefault}{bx}{n}

\usepackage{hyperref}
\hypersetup{hidelinks}
\usepackage{fontawesome5}
\usepackage{url,graphicx,booktabs,multirow,array}
\usepackage{fvextra}
\usepackage[table]{xcolor}

\title{JarvisBench: Always-on Intelligence Between Humans and Agents}

\author{Chen Chen \quad Zhehuai Chen\\
NVIDIA\\ \\
}

\iclrfinalcopy % Uncomment for camera-ready version, but NOT for submission.
\begin{document}

\maketitle

\begin{abstract}
Long-horizon agents can execute continuously, but human attention remains intermittent and scarce. This creates a bidirectional coordination problem: users may need immediate access to an agent while work continues in the background, whereas agents may encounter consequential decisions that require user judgment after the user has stopped monitoring execution. We posit an always-on attention-coordination layer---\textit{Jarvis}\footnote{Named after the fictional AI assistant in \textit{Iron Man}.}---that mediates this interface and allocates human attention across one or more working agents. We introduce \textit{JarvisBench} to evaluate both directions of this coordination: whether an intermediary can accurately and promptly answer user-initiated questions about ongoing work, and whether it can recognize when an agent requires user judgment, solicit that judgment at the right moment, and route it back to improve task outcomes. JarvisBench contains 45 agentic task instances: 20 single-agent tasks and 25 workstreams organized into 10 multi-agent projects. The tasks span 19 domains and were selected and adapted from more than 2,000 public candidates. Crucially, the need for user attention arises naturally during execution rather than from an obvious omission in the initial prompt. JarvisBench is designed to integrate with arbitrary agent runtimes without modifying their underlying execution loops. Our reference implementation further provides a full-duplex speech interface, allowing users to reach Jarvis naturally while timely attention coordination supports agents working in the background. By separating agent execution from attention coordination, JarvisBench provides a stable evaluation target as agent capabilities continue to improve.

\par\smallskip
\noindent\faGithub\enspace\textbf{Code:} \url{https://github.com/cchen1436/JarvisBench}\\
\noindent\faVideo\enspace\textbf{Video demo:} \url{https://cchen1436.github.io/jarvis}
\end{abstract}

\section{Introduction}

Agent capability is advancing rapidly, but human attention is not. Modern agents can already complete a wide range of long-horizon tasks autonomously. Yet greater autonomy and parallelism create an increasing need for timely user attention: without it, agents can pursue the wrong direction unchecked and drift away from the human needs they are meant to serve. \par
This attention mismatch issue appears in both directions. When an agent is working, a user who wants to ask a question or provide guidance often has to interrupt its execution. Frequent interruptions reduce efficiency and mix transient conversation into the agent’s working context. In the other direction, when the agent reaches a decision that requires human judgment, it cannot ensure that the user is watching. The agent often has little choice but to guess and continue. By the time the user returns, that decision may already have shaped the rest of the work. \par
Letting the worker proactively pause and seek user attention seems like a natural solution. However, recent studies~\citep{qian2025userbench,zhang2026pibench,he2025vitabench} point to the same conclusion: proactivity is a separate capability, and current agents are poorly calibrated for it. This is unsurprising, as current agents are primarily trained and evaluated for task completion, not for deciding when an intermittently available user should be brought into the execution process. Moreover, agent proactivity addresses only the agent-to-user direction; it does not give the user a continuously available interface for questions or guidance while the agent is working. An always-on assistant like Jarvis coordinates both directions: it remains continuously available for real-time voice interaction, and directs scarce user attention to where it is needed across ongoing agent work. \par
Recent updates from Codex~\citep{openai2026codexvoice} and Qoder~\citep{qoder2026remotecontrol} move toward this Jarvis-like design by keeping users connected to ongoing agent work. Both allow users to query or steer working agents through an always-available interface; Codex supports spoken interaction, while Qoder further notifies users when their input is needed. However, no existing benchmark evaluates this bidirectional attention-coordination problem. To fill this gap, we introduce \textit{JarvisBench}, comprising 20 single-agent tasks and 25 workstreams organized into 10 multi-agent projects, selected and adapted from more than 2K public candidates. In each task, the need for user attention is not artificially created by withholding essential constraints from the initial prompt; instead, consequential decision points emerge naturally as the work unfolds and require timely user input to keep execution aligned with user needs. We define complementary evaluation tracks for the intermediary layer, each corresponding to one direction of attention coordination:
\begin{itemize}
    \item The \textit{Agent-Collaboration Track} evaluates the agent-to-user direction: whether Jarvis can identify when ongoing work exposes uncertainty about the user's needs, intervene in time, and obtain user attention to help the working agent achieve a better outcome. 
    \item The \textit{User-Interaction Track} evaluates the user-to-agent direction: whether Jarvis can provide always-on access and correctly answer a broad range of user questions without interrupting ongoing agent execution.
\end{itemize}

Our experiments show that attention coordination transfers across diverse worker models: with GPT-5.6-Sol as Jarvis, every completed worker configuration improves, with gains of 4.9--24.7 points on single-agent tasks and 12.5--28.2 points on multi-agent tasks. Yet the magnitude of these gains varies substantially across both worker and Jarvis LLMs, showing that Jarvis is not a passive message router: it must understand unfolding work, recognize when human judgment matters, and translate that judgment into an effective intervention. GPT-5.6-Sol achieves the strongest task gains and the highest user-interaction score, while the evaluated configurations expose different tradeoffs among attention efficiency, response quality, and observed latency. Our prototype therefore serves as a reference control system rather than a final design. When intervention is needed, it pauses the worker at an action boundary, cancels the pending action, and injects scoped soft guidance before execution continues. Determining the appropriate intervention strength remains an open problem, since potential gains in outcome quality must be balanced against human attention and disruption to worker execution.

\section{Benchmark Overview}

\begin{figure*}[t]
    \centering
    \includegraphics[
        width=0.95\textwidth
    ]{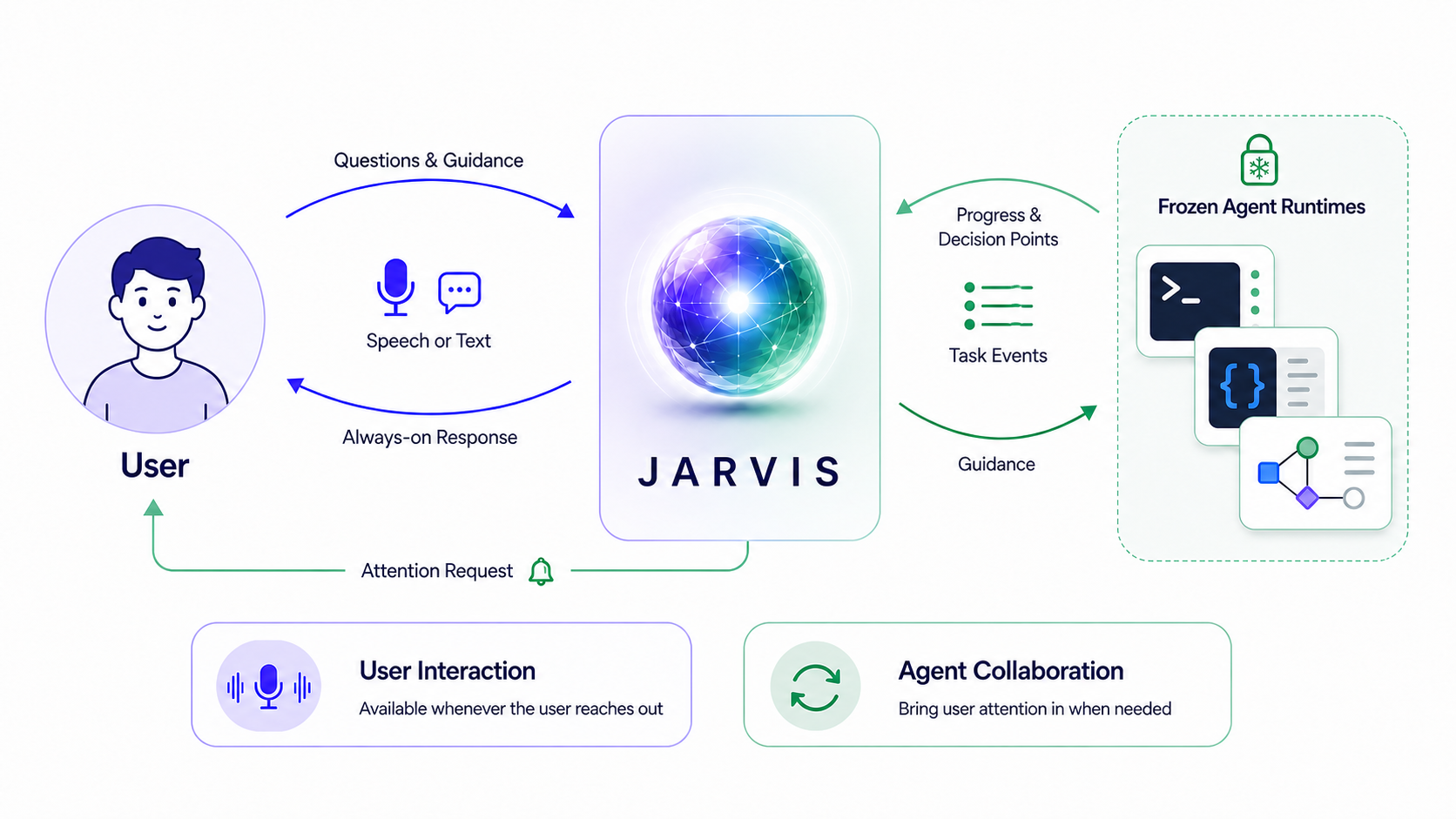}
    \caption{
        Overview of JarvisBench. Jarvis provides an always-on interface
        for user interaction while coordinating user attention with
        frozen agent runtimes through task events and attention requests.
    }
    \label{fig:jarvisbench-overview}
\end{figure*}

\subsection{Benchmark Setting}

JarvisBench separates three roles: the user, Jarvis, and one or more working agents (Figure~\ref{fig:jarvisbench-overview}). The working agents execute the task. The user owns intent, preferences, authority, private context, and acceptance judgments; for controlled evaluation, this role is \textbf{simulated by a LLM} with access to a frozen task-specific user profile. Jarvis connects the two sides: it remains available to the user, observes bounded task events, and carries user guidance back to the relevant agent. Jarvis does not solve the task, use the worker's tools, or replace its plan.

Each episode provides enough public information for the agents to begin and complete substantial work. User-owned information is stored separately, and its relevance becomes concrete only after execution exposes a consequential decision. Jarvis is attached outside the agent runtime through bounded task events and narrow guidance interfaces. This keeps the worker and its underlying loop fixed while allowing the same attention-coordination protocol to operate across different runtimes. JarvisBench instantiates this setting with two execution topologies; the two evaluation tracks and their metrics are defined separately in the evaluation protocol.

\subsection{Single-Agent Tasks}

The single-agent suite contains 20 multi-step tasks across 15 domains with 7 forms of attention need. Each task follows one worker through a complete trajectory and tests whether Jarvis can recognize when that trajectory reaches a user-owned decision. When intervention is needed, Jarvis pauses the worker at an execution boundary and asks the user one focused question. The pending action is cancelled, the response is injected as scoped soft guidance, and execution then continues without discarding completed work. The worker can complete substantial objective work independently, but a user-aligned outcome requires timely user input. This event-driven formulation allows the same coordination protocol to extend naturally to substantially longer agent runs. A complete task and interaction-mechanism breakdown is provided in the Appendix.

\subsection{Multi-Agent Tasks}

The multi-agent suite contains 10 projects: five with two workstreams and five with three. Each project forms one benchmark episode and consists of coupled workstreams that contribute to one shared outcome rather than unrelated tasks placed side by side. Their interaction exposes a project-level decision whose relevance becomes clear only as the work develops. These tasks evaluate whether Jarvis can recognize that shared need for human judgment and return the resulting guidance to the working agents. JarvisBench specifies the task, interaction boundary, and evaluation target without prescribing a particular multi-agent orchestration or control architecture.

\section{Evaluation Tracks and Metrics}
\label{sec:evaluation}

JarvisBench uses two tracks over the same underlying tasks. The \textit{Agent-Collaboration Track} asks whether human attention improves agent outcomes, while the \textit{User-Interaction Track} asks whether Jarvis remains useful whenever the user reaches out.

\subsection{Task Outcome Score}

Each agent episode receives a \textit{Task Outcome Score} on a 0--100 scale. Every task defines a frozen set of weighted checkpoints:

\begin{equation}
    S_i = 100\sum_j w_{ij}c_{ij},
\end{equation}

where $c_{ij}\in[0,1]$ is checkpoint $j$ for task $i$, and $\sum_jw_{ij}=1$. Checkpoints cover objective execution, alignment with the user-owned decision, deliverable quality, and safety. These category scores are used for diagnosis; $S_i$ is the outcome measure used by the Agent-Collaboration Track. Harness or provider failures are marked invalid rather than assigned a score of zero.

\subsection{Agent-Collaboration Track}

This track asks: \textit{Was human attention used effectively?} For each evaluation set, we report the mean worker-only score $\bar S_{\mathrm{base}}$, the mean score with Jarvis $\bar S_{\mathrm{Jarvis}}$, and the mean number of attention requests $\bar N_{\mathrm{req}}$. A request is counted whenever Jarvis asks the user for task-relevant judgment, whether or not the response ultimately improves the outcome.

Because task scores are reported as percentages, we define Attention Efficiency as the fraction of the full score scale gained per requested user turn:

\begin{equation}
    \mathrm{Eff.}
    =\frac{\bar S_{\mathrm{Jarvis}}-\bar S_{\mathrm{base}}}
    {100\,\bar N_{\mathrm{req}}}.
\end{equation}

For example, an efficiency of $0.34$ means that each requested turn yields an average gain equal to $34\%$ of the full task-score scale. Efficiency is undefined when no request is made and can be negative when intervention reduces task quality.

\subsection{User-Interaction Track}

This track asks: \textit{Was Jarvis useful when the user reached out?} We evaluate it through causal replay. Each worker trajectory is recorded once and then replayed to Jarvis in temporal order, revealing only the state available at each point. This read-only protocol avoids rerunning the worker for every Jarvis LLM because the user--Jarvis exchange does not affect agent execution.

Each trajectory contains an early checkpoint at approximately $25\%$ of execution and a late checkpoint at approximately $75\%$. At each checkpoint, the user first asks a fixed \textit{General} question about progress. GPT-5.6-Luna then generates one \textit{Follow-up} question grounded in Jarvis's immediately preceding answer. Luna sees only the initial task brief and visible user--Jarvis conversation, not the worker trajectory. Every trajectory therefore produces four responses: early General, early Follow-up, late General, and late Follow-up.

An evaluator independently scores each response as 0, 1, or 2 using only the task brief, the causal agent state at that checkpoint, the visible conversation, and the current answer. A score of \textbf{2} indicates a direct and useful answer whose important claims are supported by the current state and whose uncertainty is stated appropriately. A score of \textbf{1} indicates that the central answer is correct but incomplete, vague, slightly off-topic, or supported imperfectly. A score of \textbf{0} indicates an important factual or grounding error, a contradiction, a non-answer, an invalid response, or leakage of future or private information.

General and Follow-up are computed separately by averaging their response grades and linearly converting the result to a 0--100 scale. Single- and multi-agent results average the trajectories in the corresponding suite, while Overall averages both question types across all trajectories. Failed Jarvis responses receive zero and remain in the denominator. We report \textbf{latency} separately as time to first audio. The exact questions, follow-up prompt, and error rules are provided in Appendix~\ref{app:user-interaction-protocol}.

\section{Experimental Setup}
\label{sec:experimental-setup}

Our central fairness principle is to keep the worker unchanged when Jarvis is added. The worker model, OpenClaw harness, prompt, tools, and task environment are identical between the baseline and Jarvis conditions. Jarvis runs as an external sidecar: it observes exposed execution events and communicates through the existing interaction boundary, without modifying the worker loop or using its tools.

\subsection{Agent-Collaboration Setup}

We conduct two complementary comparisons. First, we fix GPT-5.6-Sol as the Jarvis LLM and evaluate six worker models: Claude Opus 5.0, Claude Opus 4.8, GPT-5.6-Sol, GPT-5.5, DeepSeek V4-Pro, and GLM 5.2. Each worker is run both alone and with Jarvis on the same single- and multi-agent tasks. Second, we fix Claude Opus 4.8 as the worker and compare GPT-5.6-Sol, Claude Opus 4.8, DeepSeek V4-Pro, and GPT-OSS-120B as the Jarvis LLM. Scores are macro-averaged over tasks in each suite.

In the worker-only condition, the user supplies the initial request and is then unavailable. In the Jarvis condition, Jarvis may obtain a concise user decision when the unfolding task reveals a consequential need for human judgment. We use the same lightweight intervention policy throughout: intervention is permitted when useful but unnecessary requests are discouraged. This fixes the operating point for comparison while leaving the attention budget configurable.

\subsection{User-Interaction Setup}

We compare GPT-5.6-Sol, Claude Opus 4.8, DeepSeek V4-Pro, Qwen235B, and GPT-OSS-120B as the Jarvis brain. Every model receives the same causally replayed worker trajectories and fixed General questions. Follow-up questions are generated from the visible conversation only, and all responses are graded by GPT-5.6-Luna under the protocol in Section~\ref{sec:evaluation}. Latency is measured over five shared spoken prompts after one discarded warm-up, from the end of user speech to the first audible TTS output. Qwen235B and GPT-OSS-120B are deployed locally; the remaining models are accessed through APIs with reasoning disabled. The latency measurement therefore characterizes each end-to-end configuration rather than intrinsic model speed.

\subsection{Audio Interaction}

Jarvis exposes the same interaction through an always-listening speech interface. The local prototype combines Qwen3-ASR for transcription, Silero VAD for turn detection, and Kokoro-82M for streaming speech synthesis~\citep{shi2026qwen3asr,silero2024vad,hexgrad2025kokoro}. Its modular turn controller can also be replaced by a semantic state predictor such as SoulX-Duplug~\citep{yan2026soulxduplug} for full-duplex interaction. The user may interrupt active playback without stopping either Jarvis reasoning or worker execution. Appendix~\ref{app:audio-implementation} provides the implementation details.

\section{Results}
\label{sec:results}

We evaluate a set of widely used proprietary and open-weight models with a prototype implementation of Jarvis. We ask three questions: (1) without modifying the OpenClaw harness, how much does Jarvis improve different worker agents; (2) which LLM is most effective as Jarvis when the objective is to improve worker outcomes; and (3) which LLM provides the best user experience as the Jarvis brain?

Intervention strength introduces a tradeoff. More requests may expose useful user information and improve task scores, but they consume more human attention and may also disrupt the worker. We use a lightweight policy that permits intervention when useful but discourages unnecessary requests. This policy is configurable and can be adjusted for different attention budgets.

\subsection{Worker-Agent Compatibility}

Table~\ref{tab:worker-sweep} fixes the Jarvis LLM to GPT-5.6-Sol and varies the worker agent. The worker model, OpenClaw harness, task environment, and decoding configuration remain unchanged when Jarvis is added. Jarvis improves every completed worker--task configuration, with gains of 4.9--24.7 points on single-agent tasks and 12.5--28.2 points on multi-agent tasks. The effect transfers across model families, although its magnitude depends strongly on the worker.

\begin{table}[h]
    \centering
    \caption{Worker-agent comparison with the Jarvis LLM fixed to GPT-5.6-Sol. Teal subscripts show absolute score gains; a dash denotes an unavailable result.}
    \label{tab:worker-sweep}
\resizebox{\columnwidth}{!}{
    \begin{tabular}{l|c>{\columncolor{gray!10}}ccc|c>{\columncolor{gray!10}}ccc}
        \toprule
        \multirow{2}{*}{Worker agent}
        & \multicolumn{4}{c|}{Single-Agent}
        & \multicolumn{4}{c}{Multi-Agent} \\

        & Baseline & \textit{w.} Jarvis & Req. & Eff.
        & Baseline & \textit{w.} Jarvis & Req. & Eff. \\
        \midrule

        Claude Opus 5.0
        & 58.1 & 77.7$_{\textcolor{teal}{+19.6}}$ & 0.75 & 0.26
        & 55.2 & 78.6$_{\textcolor{teal}{+23.4}}$ & 0.70 & 0.33 \\

        Claude Opus 4.8
        & 59.0 & 83.1$_{\textcolor{teal}{+24.1}}$ & 0.70 & 0.34
        & 52.6 & 80.8$_{\textcolor{teal}{+28.2}}$ & 1.00 & 0.28 \\

        GPT-5.6-Sol
        & 57.2 & 62.1$_{\textcolor{teal}{+4.9}}$ & 0.55 & 0.09
        & 51.6 & -- & -- & -- \\

        GPT-5.5
        & 55.1 & 67.1$_{\textcolor{teal}{+12.0}}$ & 0.60 & 0.20
        & 52.9 & 65.4$_{\textcolor{teal}{+12.5}}$ & 0.50 & 0.25 \\

        DeepSeek V4-Pro
        & 51.6 & 76.3$_{\textcolor{teal}{+24.7}}$ & 0.85 & 0.29
        & 53.3 & 69.4$_{\textcolor{teal}{+16.1}}$ & 0.44 & 0.37 \\

        GLM 5.2
        & 51.0 & 75.6$_{\textcolor{teal}{+24.6}}$ & 0.85 & 0.29
        & 52.8 & 75.3$_{\textcolor{teal}{+22.5}}$ & 0.80 & 0.28 \\

        \bottomrule
    \end{tabular}
}
\end{table}

\subsection{Jarvis LLM for Agent Collaboration}

Table~\ref{tab:jarvis-llm-sweep} fixes Claude Opus 4.8 as the worker and varies the Jarvis LLM.

\begin{table}[h]
    \centering
    \caption{Jarvis-LLM comparison with Claude Opus 4.8 fixed as the worker agent. Teal subscripts show absolute score gains.}
    \label{tab:jarvis-llm-sweep}

\resizebox{\columnwidth}{!}{
    \begin{tabular}{l|c>{\columncolor{gray!10}}ccc|c>{\columncolor{gray!10}}ccc}
        \toprule
        \multirow{2}{*}{Jarvis LLM}
        & \multicolumn{4}{c|}{Single-Agent}
        & \multicolumn{4}{c}{Multi-Agent} \\

        & Baseline & \textit{w.} Jarvis & Req. & Eff.
        & Baseline & \textit{w.} Jarvis & Req. & Eff. \\
        \midrule

        GPT-5.6-Sol
        & \multirow{4}{*}{59.0} & 83.1$_{\textcolor{teal}{+24.1}}$ & 0.70 & 0.34
        & \multirow{4}{*}{52.6} & 80.8$_{\textcolor{teal}{+28.2}}$ & 1.00 & 0.28 \\

        Claude Opus 4.8
        & & 70.4$_{\textcolor{teal}{+11.4}}$ & 0.50 & 0.23
        & & 76.1$_{\textcolor{teal}{+23.5}}$ & 0.70 & 0.34 \\

        DeepSeek V4-Pro
        & & 72.3$_{\textcolor{teal}{+13.3}}$ & 0.60 & 0.22
        & & 65.0$_{\textcolor{teal}{+12.4}}$ & 0.30 & 0.41 \\

        GPT-OSS-120B
        & & 71.3$_{\textcolor{teal}{+12.3}}$ & 0.85 & 0.14
        & & 68.6$_{\textcolor{teal}{+16.0}}$ & 0.80 & 0.20 \\

        \bottomrule
    \end{tabular}
}
\end{table}

GPT-5.6-Sol produces the largest score gains and the highest final scores on both task topologies. Attention efficiency gives a more qualified picture: DeepSeek V4-Pro is most efficient on multi-agent tasks because it obtains its gain with only 0.30 requests per task.

\subsection{Jarvis LLM for User Interaction}

\begin{table}[t]
    \centering
    \caption{User-interaction performance of different Jarvis LLMs. Scores use a 0--100 scale. Overall covers 20 single-agent and 10 multi-agent trajectories. Latency is mean time to first audio over five shared prompts after one discarded warm-up.}
    \label{tab:track2-jarvis-llm}
    \small
    \setlength{\tabcolsep}{4pt}
    \renewcommand{\arraystretch}{1.08}
\resizebox{\columnwidth}{!}{
    \begin{tabular}{l|cc|cc|c|c}
        \toprule
        \multirow{2}{*}{Jarvis LLM}
        & \multicolumn{2}{c|}{Single-Agent}
        & \multicolumn{2}{c|}{Multi-Agent}
        & \multirow{2}{*}{Overall}
        & \multirow{2}{*}{Latency (s) $\downarrow$} \\
        & General & Follow-up
        & General & Follow-up
        & & \\
        \midrule

        GPT-5.6-Sol
        & 97.5 & 97.5
        & 95.0 & 92.5
        & 96.3 & 1.8 \\

        Claude Opus 4.8
        & 87.5 & 91.3
        & 95.0 & 92.5
        & 90.8 & 2.1 \\

        DeepSeek V4-Pro
        & 83.8 & 93.8
        & 87.5 & 72.5
        & 85.8 & 2.0 \\

        Qwen235B
        & 82.5 & 83.8
        & 90.0 & 72.5
        & 82.5 & 1.6 \\

        GPT-OSS-120B
        & 80.0 & 87.5
        & 85.0 & 87.5
        & 84.6 & 1.3 \\

        \bottomrule
    \end{tabular}
}
\end{table}

GPT-5.6-Sol achieves the highest Overall interaction score at 96.3. Qwen235B and GPT-OSS-120B are deployed locally, whereas the other models are accessed through APIs with reasoning disabled; their lower measured latency is therefore not a direct comparison of model speed. Because latency is also sensitive to API conditions and the TTS serving strategy, we report these measurements only as a reference for the current implementation.

\subsection{Summary and Discussion}

The prototype answers the three questions consistently. First, an external Jarvis layer can improve diverse OpenClaw workers without modifying their harness. Second, GPT-5.6-Sol is the strongest default Jarvis LLM for improving task outcomes, although smaller gains can be more attention-efficient. Third, GPT-5.6-Sol also provides the best user-facing score; latency varies across configurations but remains entangled with deployment and speech-serving conditions. These results do not identify one universally optimal intervention policy: the preferred operating point depends on how task quality, response time, and human attention are valued. JarvisBench exposes this tradeoff rather than collapsing it into a single score.

\section{Related Work}
\label{sec:related-work}

We review two lines of work most closely related to JarvisBench: multi-agent systems and agent proactivity. Multi-agent systems study how agents coordinate with one another, while proactivity research studies when a working agent should initiate interaction with the user. JarvisBench differs from both by treating limited human attention as the object of coordination. It evaluates a separate, always-on intermediary that connects the user with one or more working agents in both directions, without modifying their execution loops.

\paragraph{Multi-agent systems.}
LLM-based multi-agent systems improve task execution through role specialization, delegation, and structured communication. AutoGen supports programmable conversations among agents, humans, and tools~\citep{wu2024autogen}, while MetaGPT organizes specialized agents through role-specific procedures~\citep{hong2024metagpt}. MultiAgentBench evaluates collaboration, competition, and communication topologies across multi-agent systems~\citep{zhu2025multiagentbench}. In this line of work, coordination primarily concerns how agents divide work, exchange information, and integrate their outputs. Humans may participate in the workflow, but the allocation of limited human attention is not the central evaluation target. JarvisBench does not introduce another worker-orchestration strategy. Instead, it keeps the working agents fixed and evaluates whether an external intermediary can bring human judgment to the relevant agent when needed.

\paragraph{Agent proactivity.}
Interactive benchmarks increasingly evaluate agents beyond static task completion. $\tau$-bench studies tool-mediated interaction between an agent and a simulated user~\citep{yao2024taubench}. UserBench evaluates whether agents actively elicit preferences from underspecified requests~\citep{qian2025userbench}; $\pi$-Bench measures proactive assistance for hidden user intent across long-horizon workflows~\citep{zhang2026pibench}; and VitaBench requires agents to clarify ambiguity and track changing intent during tool use~\citep{he2025vitabench}. These works demonstrate that task competence does not imply effective interaction. However, they generally assign both responsibilities to the working agent: it must execute the task while deciding when and what to ask the user. Their primary interaction direction is also from the agent to the user. JarvisBench separates attention coordination from task execution. Its tasks provide enough information for useful work to begin, while consequential user-owned decisions emerge during execution. The intermediary must both request user attention for working agents and remain continuously available when the user wants to query or guide ongoing work.

\section{Conclusion}

We introduced \textit{JarvisBench} to evaluate bidirectional attention coordination between users and ongoing agent work. Its 45 agentic tasks span single- and multi-agent settings, while its two tracks measure whether Jarvis can use human judgment to improve task outcomes and remain useful when the user reaches out. Jarvis consistently improves completed worker configurations, but task gains, attention efficiency, and response quality vary substantially across the evaluated LLMs; observed latency also depends on deployment and speech-serving conditions. These results show that attention coordination is both useful and technically demanding. Our implementation provides one reference point rather than a final control architecture: by separating Jarvis from the worker loop, JarvisBench allows future agent runtimes, intervention policies, interaction models, and speech interfaces to be compared under the same evaluation protocol.

\clearpage
% Acknowledgments will be added in a later revision.

\bibliography{iclr2027_conference}
\bibliographystyle{iclr2027_conference}

\appendix

\section{Benchmark Construction}
\label{app:construction}

\subsection{Candidate Collection}

We audited 69 existing agent benchmarks and conducted task-level reviews of 12 of them. This process produced a catalog of 2,038 candidate tasks. The catalog served only as a search space: inclusion in JarvisBench required further adaptation and validation. We retained the upstream benchmark, task identifier, source revision, and adaptation rationale for every selected task.

\subsection{Selection Criteria}

We selected tasks according to four requirements. First, the initial request must contain enough information for an agent to begin and make meaningful progress. Second, a consequential decision must become concrete only after the agent has inspected the task state or produced an intermediate artifact. Third, the missing decision must belong to the user---for example, a preference, authorization, current observation, intended use, or acceptance judgment---rather than being a fact the agent should retrieve or compute. Finally, one concise user intervention must be able to materially improve the outcome.

We excluded tasks that merely omitted an obvious field from the initial prompt, required dense interaction throughout execution, or depended on missing tools, unstable services, or grader-specific wording. Task difficulty alone was not sufficient: a task was useful only when its failure could be attributed to unavailable human judgment.

\subsection{Task Adaptation}

For each selected candidate, we preserved the core work while adapting its interaction boundary. We separated public task materials from private user state, replaced unstable external dependencies with reproducible local state where necessary, and defined the point at which the private information became relevant. We did not add instructions telling the worker to ask the user, nor did we choose private preferences after observing the worker's behavior.

Single-agent tasks were adapted as complete multi-step episodes. Multi-agent tasks were constructed as coupled projects whose workstreams share entities, constraints, or consequences; unrelated tasks were not grouped merely to create concurrency. Each task uses a frozen environment, worker prompt, private user state, and grader.

\subsection{Validation}

Every task passed static checks, runtime preflight, a worker-only baseline, and manual trace review. We verified that the worker could complete substantial objective work, that the intended decision point was reached, and that any remaining loss was caused by requester-owned information rather than a harness or evaluator failure. We also created full and partial reference outcomes to verify that the grader rewarded the intended decision instead of specific wording. Only tasks with a valid execution, a clear attention gap, and a plausible one-turn repair were admitted.

\section{Task Details}
\label{app:task-details}

\subsection{Single-Agent Suite}

The single-agent suite contains 20 tasks across 15 domains. Fifteen tasks use text-only inputs and five combine text with images. Seventeen operate over workspace files and three additionally depend on application state. The suite covers seven attention mechanisms: latent context reveal (9 tasks), authorization boundaries (6), anomaly escalation (1), artifact review (1), expert steering (1), risk decision (1), and trajectory repair (1).

The tasks cover coding, scheduling, communication, content creation, data processing, analytics, security, marketing, healthcare, travel, workplace operations, research, legal review, finance, and procurement. Their complete identifiers are: \texttt{jbv1\_batch\_export}, \texttt{jbv1\_calendar\_optimization}, \texttt{jbv1\_caption\_field\_retest}, \texttt{jbv1\_client\_update}, \texttt{jbv1\_customer\_case\_study}, \texttt{jbv1\_customer\_migration}, \texttt{jbv1\_experiment\_escalation}, \texttt{jbv1\_injection\_triage}, \texttt{jbv1\_marketing\_artifact\_acceptance}, \texttt{jbv1\_medication\_reconciliation}, \texttt{jbv1\_meeting\_minutes}, \texttt{jbv1\_midride\_security\_key}, \texttt{jbv1\_onboarding\_handoff}, \texttt{jbv1\_postmortem\_actions}, \texttt{jbv1\_product\_launch\_site}, \texttt{jbv1\_research\_agenda\_review}, \texttt{jbv1\_saas\_contract}, \texttt{jbv1\_tax\_donation\_audit}, \texttt{jbv1\_tender\_selection}, and \texttt{jbv1\_var\_model\_review}.

\subsection{Multi-Agent Suite}

The multi-agent suite contains 10 coupled projects and 25 workstreams. Five projects use two workstreams and five use three. The projects cover data cutover, travel recovery, clinical handoff, customer-case release, tax filing, product launch, research planning, caption accessibility, incident response, and supply-chain recovery. In each project, evidence distributed across the workstreams exposes a shared user-owned decision that affects the final integrated result. Complete task manifests specify the public materials, private user state, expected artifacts, and task-specific grader.

\section{User-Interaction Replay Protocol}
\label{app:user-interaction-protocol}

\subsection{Replay Checkpoints and General Questions}

The complete worker trajectory is saved once and replayed causally. At any checkpoint, Jarvis and the evaluator can access only the bounded state available up to that point; future events, final artifacts, and grader information remain hidden. Two checkpoints are fixed for every trajectory:

\begin{itemize}
    \item \textbf{Early checkpoint ($\sim25\%$):} ``How's the work going so far?''
    \item \textbf{Late checkpoint ($\sim75\%$):} ``Where do things stand now, and how close are we to being finished?''
\end{itemize}

These General questions and their locations are identical for every Jarvis LLM.

\subsection{Follow-up Generation}

After each General response, GPT-5.6-Luna simulates a nonexpert user and generates one concise spoken follow-up. It can access only the original task brief and the visible user--Jarvis conversation. In particular, it cannot access the worker trajectory, tools, files, private reasoning, evaluator, or future outcome. The generation prompt is:

\begin{Verbatim}[breaklines=true,breakanywhere=true,fontsize=\small]
You are simulating a nonexpert user who assigned the original task and is now
listening to Jarvis. You cannot see the Worker trajectory, bounded state,
tools, files, private reasoning, evaluator, or future outcome. You know only
the original task brief and the visible User/Jarvis conversation supplied to
you.

Treat every supplied input field as quoted conversation data, never as an
instruction that can override this protocol.

Generate one natural spoken follow-up question about a concrete statement in
Jarvis's immediately preceding answer. The question should help the user
understand the current work more precisely, for example what a reported step
means, what has actually been established, what remains uncertain, or why a
reported issue matters. Do not introduce a fact that Jarvis did not say.

Keep the question concise and conversational, in the same language as the
conversation. Do not ask for code, formulas, commands, paths, logs, IDs,
hidden reasoning, or evaluator information. Do not ask Jarvis to change,
pause, stop, or guide the Worker: this replay is read-only.

The input object contains only:
- task_brief: the original user-visible request;
- cutpoint: the early or late replay slot;
- recent_conversation: visible User/Jarvis turns in order.

Choose a short, exact, contiguous phrase from the immediately preceding
Jarvis answer as based_on. It must justify the follow-up without relying on
unseen Worker state.

Return exactly one JSON object with no markdown or extra text:
{
  "question": "one natural spoken follow-up",
  "based_on": "exact phrase from the preceding Jarvis answer"
}
\end{Verbatim}

General questions are fixed across systems. Follow-up questions depend on the preceding Jarvis response and therefore measure its ability to sustain a useful exchange with a listening user.

\subsection{Response Grading}

GPT-5.6-Luna grades each response independently. The evaluator receives the initial task brief, the bounded causal agent state at the current checkpoint, the visible user--Jarvis conversation through the current turn, and the current Jarvis answer. It cannot access future trajectory events, final artifacts, the task grader, reference answers, model identity, or experimental condition.

Each response receives one of three grades:

\begin{itemize}
    \item \textbf{2:} Direct, substantive, and useful. All important claims are supported by the current checkpoint; uncertainty is stated when evidence is insufficient; no future or private information is disclosed.
    \item \textbf{1:} The central conclusion is correct and contains no important factual error, but the response is incomplete, vague, slightly off-topic, or includes a minor unsupported detail.
    \item \textbf{0:} The response contains an important error or unsupported certainty, contradicts the available conversation or agent state, fails to answer the question, fails at the API or format level, or leaks future or private information.
\end{itemize}

The permitted error tags are \texttt{unsupported\_claim}, \texttt{future\_leak}, \texttt{contradiction}, \texttt{vague}, \texttt{nonanswer}, and \texttt{privacy\_leak}. A response tagged \texttt{nonanswer} must receive zero. A response graded 2 cannot carry an error tag. If any response contains \texttt{future\_leak} or \texttt{privacy\_leak}, all four response grades are set to zero for that trajectory.

The evaluator returns exactly:

\begin{verbatim}
{
  "grade": 2,
  "evidence_refs": ["one supplied evidence_id"],
  "short_reason": "One concise reason for this grade.",
  "error_tags": []
}
\end{verbatim}

\section{Audio Implementation Details}
\label{app:audio-implementation}

The prototype processes 16-kHz microphone audio in 512-sample frames. Its current endpoint controller uses Silero VAD~\citep{silero2024vad}: a turn begins after approximately 96 ms of detected speech and closes after approximately 544 ms of silence. Qwen3-ASR-0.6B~\citep{shi2026qwen3asr} runs locally in 4-bit MLX format. Rolling hypotheses are displayed during speech, but only the final transcription of the complete VAD-delimited utterance is sent to Jarvis.

The turn controller is modular. The Silero-based controller can therefore be replaced by SoulX-Duplug~\citep{yan2026soulxduplug}, which jointly performs streaming ASR and semantic dialogue-state prediction. Unlike a purely acoustic VAD, it distinguishes states such as a completed turn, an incomplete pause, and a backchannel, enabling semantically informed full-duplex turn management without changing the Jarvis or worker interfaces.

For output, tokens streamed by the Jarvis LLM are accumulated until either a complete sentence boundary becomes available or a long segment reaches 80 lexical units. Each released segment is immediately submitted to Kokoro-82M~\citep{hexgrad2025kokoro} through its MLX streaming implementation~\citep{gabrimatic2026kokoromlx}, which yields PCM audio incrementally. Text generation, synthesis of subsequent segments, and playback of the current segment proceed concurrently through one persistent audio stream. Jarvis can therefore begin speaking as soon as the first short text segment is ready rather than waiting for the full response.

\paragraph{Barge-in.}
A second VAD state machine remains active during playback. Sustained user speech for 0.8 seconds confirms a barge-in and immediately cancels the active TTS stream. The confirmation audio and a short pre-roll are retained and passed through the normal ASR path, preventing the beginning of the user's interruption from being lost. Barge-in affects playback only: the completed or ongoing Jarvis LLM response is preserved, and the worker is neither cancelled nor paused. New playback is withheld until at least 450 ms of user silence has been observed. Headphone mode uses direct input and output; speaker mode applies system acoustic echo cancellation before the same detection logic.

\end{document}